\documentclass[conference]{IEEEtran}
\IEEEoverridecommandlockouts
\usepackage{cite}
\usepackage{amsmath,amssymb,amsfonts}
\usepackage{algorithmic}
\usepackage{graphicx}
\usepackage{textcomp} 
\usepackage{array}
\usepackage{booktabs}
\usepackage{xcolor} 
\usepackage{tikz} 
\usepackage{float}  
\usepackage[hidelinks]{hyperref}  

\usepackage{hyperref} 
\usepackage[numbers]{natbib}
\def\BibTeX{{\rm B\kern-.05em{\sc i\kern-.025em b}\kern-.08em
    T\kern-.1667em\lower.7ex\hbox{E}\kern-.125emX}}
\begin{document}

\newcommand{\fw}{\text{future work}}
\newcommand{\futurework}{\text{future work}}

\title{FutureGen: A RAG-based Approach to Generate the Future Work of Scientific
Article \\
}

\author{
\IEEEauthorblockN{Ibrahim Al Azher}
\IEEEauthorblockA{\textit{Dept of Computer Science} \\
\textit{Northern Illinois University}\\
Dekalb, IL, USA \\
iazher1@niu.edu}
\and 
\IEEEauthorblockN{Miftahul Jannat Mokarrama}
\IEEEauthorblockA{\textit{Dept of Computer Science} \\
\textit{Northern Illinois University}\\
Dekalb, IL, USA \\
mmokarrama1@niu.edu}
\and
\IEEEauthorblockN{Zhishuai Guo}
\IEEEauthorblockA{\textit{Dept of Computer Science} \\
\textit{Northern Illinois University}\\
Dekalb, IL, USA \\
zguo@niu.edu} 
\and 

\IEEEauthorblockN{Sagnik Ray Choudhury}
\IEEEauthorblockA{\textit{Dept of Computer Science} \\
\textit{University of North Texas}\\
Denton, TX, USA \\
sagnik.raychoudhury@unt.edu}
\and
\IEEEauthorblockN{Hamed Alhoori}
\IEEEauthorblockA{\textit{Dept of Computer Science} \\
\textit{Northern Illinois University}\\
Dekalb, IL, USA \\
alhoori@niu.edu}
}

\maketitle
\begin{abstract}
The Future Work section of a scientific article outlines potential research directions by identifying gaps and limitations of a current study. This section serves as a valuable resource for early-career researchers seeking unexplored areas and experienced researchers looking for new projects or collaborations. In this study, we generate future work suggestions from a scientific article. 
To enrich the generation process with broader insights and reduce the chance of missing important research directions, we use context from related papers using RAG. We experimented with various Large Language Models (LLMs) integrated into Retrieval-Augmented Generation (RAG). We incorporate an LLM feedback mechanism to enhance the quality of the generated content and introduce an LLM-as-a-judge framework for robust evaluation, assessing key aspects such as novelty, hallucination, and feasibility. Our results demonstrate that the RAG-based approach using GPT-4o mini, combined with an LLM feedback mechanism, outperforms other methods based on both qualitative and quantitative evaluations. Moreover, we conduct a human evaluation to assess the LLM as an extractor, generator, and feedback provider. 
The code is available at \url{https://github.com/IbrahimAlAzhar/FutureWorkGeneration}, and the dataset repository is available at \url{https://huggingface.co/datasets/iaadlab/FutureGen}
\end{abstract}


\section{Introduction} 
The Future Work section in a scientific article plays a crucial role in amplifying a study’s impact by demonstrating foresight and highlighting the broader implications of research \cite{b1,b2}. It serves as a catalyst for further exploration, interdisciplinary collaboration, and new ideas, transforming a single study into a foundation for future advancements \cite{b4}. Beyond academia, Future Work insights benefit policymakers and funding agencies by identifying emerging research directions and prioritizing areas for strategic resource allocation \cite{b5,b6}. A well-constructed Future Work section serves both methodological and practical purposes. For example, it encourages researchers to critically reflect on their study’s limitations, fostering higher-quality subsequent research while also streamlining the peer review process by clarifying the authors’ awareness of challenges and next steps \cite{b8}. Recognizing the limitations of a study provides clear signals for further exploration, bridging current findings with future advancements \cite{b9}. Additionally, generating sugesstive Future Work helps researchers align with current priorities, uncover unexplored gaps, and avoid redundancy. Furthermore, understanding long-term research trajectories allows early-career researchers to identify high-impact topics and strategically position their contributions in advancing scientific progress \cite{b10}. 


Author-written Future Work sections are, by nature, often speculative and may be expressed in broad or ambiguous terms \cite{b11}. While this exploratory character is valuable for signaling open directions, it can make such sections difficult to locate, interpret, and systematically compare across papers. The level of detail also varies across research communities and venues, where factors such as page limits, reviewer expectations, and time constraints shape how future directions are articulated. Moreover, even when such sections are included, they may receive limited attention post-publication, as readers often focus primarily on the main contributions \cite{b50}. To address these challenges, we generate Future Work automatically from each paper and evaluate it against both the author-mentioned future work and long-term goals extracted from OpenReview peer reviews. By combining these peer-reviewed objectives with author-written statements, our approach aims to construct a more comprehensive and reliable ground truth for evaluating future work generation.

Advances in artificial intelligence (AI) offer transformative potential for addressing this gap. Unlike traditional methods, AI can systematically synthesize research trajectories, uncover latent connections, and propose novel directions that align with emerging trends \cite{b12}. For example, Si et al. \cite{b13} demonstrate that LLM-generated ideas are more novel than those generated by human experts. 
However, current applications of AI like ChatGPT risk homogenizing outputs and reducing individual creativity \cite{b14, b15}. Standard LLMs may generate overgeneralized, irrelevant, or fabricated Future Work directions. To address these challenges, this work utilizes an LLM to suggest Future Work, enhances its output using LLM-based feedback, and incorporates RAG using cross-domain insights.

Evaluation of AI-generated Future Work sentences is challenging. Traditional Natural Language Processing (NLP) evaluation metrics that rely on n-gram text overlaps or semantic similarity, such as ROUGE \cite{b16}, BLEU \cite{b17}, and BERTScore \cite{b18} to compare the generated text with references, fail to fully capture the nuances of this generation process.
To address this issue, we incorporate LLM-based evaluation alongside NLP-based evaluations, providing human-like assessments with explanations. If the generated Future Work does not meet quality thresholds, we refine it using an LLM-driven feedback loop, improving its alignment with the context. We mitigate issues such as vagueness and redundancy observed in prior AI-driven ideation tools by integrating iterative large language model (LLM) feedback. 

The goal of this research is not to replace authors in writing their own Future Work sections, but to develop a framework that can automatically extract, generate, and evaluate Future Work statements from scientific papers. This serves two complementary purposes: (1) enabling authors and researchers to obtain higher-quality and more diverse suggestions for Future Work in their own studies, and (2) allowing the broader community to systematically identify and analyze research directions across large collections of papers. To this end, we propose an LLM-based framework that integrates Retrieval-Augmented Generation (RAG), iterative LLM feedback, and LLM-as-a-judge evaluation.

Moreover, in our study, we'll be addressing these research questions: \\ 
\textbf{RQ1:} How reliable is an LLM in extracting future work statements compared to traditional tool-based methods? \\ 
\textbf{RQ2:} How does input context selection (top-3 vs. all sections) impact the quality of LLM-generated future work, and how does RAG influence these results? \\ 
\textbf{RQ3:} How does expanding ground truth with peer-review goals and incorporating LLM feedback impact the performance and quality of LLM-generated future work? \\

Our contributions can be summarized as follows: 

\textbf{Dataset.} We created a dataset of Future Work sentences from nearly 4,000 papers collected from the ACL conference from 2023 and 2024 and 1000 papers from NeurIPS. Papers often do not have exclusive Future Work sections (they are combined with general conclusions or limitations), therefore, we use LLMs to extract relevant sentences. We also validated this extraction process on a random sample of the dataset with human annotators, ensuring that the LLM accurately identifies future-work content. \\  
\textbf{Future Work Generation.} To address RQ2, we prompted LLMs to generate future work suggestions using two input configurations: the top-3 most relevant sections and the full content (excluding the ground truth future work). Also, we developed an RAG-based system that augments the input with relevant content from related papers, enriching the generation process and improving the depth and relevance of the suggested future directions. \\  
\textbf{LLM Feedback and Judgment Framework.} To address RQ3, we incorporated long-term research goals from OpenReview to enrich the ground truth, applied LLM feedback to iteratively refine the generated future work, and used an LLM as a judge to evaluate quality—moving beyond traditional NLP-based metrics to assess dimensions such as novelty, feasibility, and hallucination. \\ 

\section{Related Work}
Recent advancements in NLP and LLMs have enabled the automatic extraction and generation of various sections of scientific articles, including abstracts \cite{b20}, methodologies \cite{b21}, charts and graphs \cite{b24}, and limitations \cite{b53, b22, b23}. Within the domain of Future Work, prior studies have focused on tasks such as extraction \cite{b25}, classification \cite{b26}, identifying creative topics \cite{b27}, thematic categorization \cite{b28}, and trend prediction \cite{b29}. For example, a BERT-based model has been used to annotate ``future research” sentences, enabling clustering \cite{b30} and impact analysis \cite{b31}. Other efforts have integrated retrieval-augmented methods for idea generation \cite{b32} and trend forecasting \cite{b51}. However, most of these approaches emphasize identifying or categorizing Future Work rather than generating actionable suggestions. Our work addresses this gap by leveraging LLMs to synthesize suggestive Future Work informed by both paper content and peer-review insights.

Beyond Future Work, the application of LLMs in scientific discovery has gained significant attention. In the biomedical domain, LLMs have been used to generate new discoveries \cite{b33}, while social science studies employed LLM-based agents to automatically propose and test hypotheses \cite{b34}. Similarly, probabilistic models have been explored for hypothesis generation using reward functions \cite{b35}. Statistical tests further show that LLM-generated research ideas can exhibit greater novelty than human-generated ones when comparing outputs against topics from recent conferences \cite{b36, b37}. These findings underscore the potential of LLMs for ideation, though their application to automated Future Work generation remains underexplored.

To enhance factual grounding, Retrieval-Augmented Generation (RAG) has emerged as a promising strategy. RAG improves generation by retrieving relevant passages from external corpora and conditioning the model’s output on this evidence, thereby reducing hallucination and improving relevance \cite{b51}. In our framework, we employ RAG to generate Future Work suggestions that align more closely with both the source paper and peer-review feedback.

A further challenge lies in improving the quality of LLM-generated content. Prior work has shown that human feedback can enhance outputs \cite{b38}, while self-refinement methods allow LLMs to identify and correct their own mistakes without human intervention \cite{b39}. For example, iterative fine-tuning approaches \cite{b40}, in-context prompt criticism \cite{b41}, and feedback-driven hypothesis generation \cite{b42, b43} all highlight the potential of feedback mechanisms. Inspired by these approaches, we incorporate LLM feedback loops to refine generated Future Work and use the LLM itself as a judge, thereby improving fluency, coherence, and originality without additional training.

Finally, the evaluation of LLM-generated text has been widely studied. Research comparing human and LLM judgments \cite{b45, b46} shows that LLM-based evaluations align well with human annotators. Building on these insights, our framework combines NLP-based metrics (e.g., ROUGE, BLEU, BERTScore) with LLM-based evaluation dimensions such as novelty, feasibility, coherence, and hallucination.

In summary, while prior research has advanced extraction, classification, and trend prediction of Future Work, as well as LLM-driven scientific discovery, the suggestive generation of Future Work remains an open challenge. Our work addresses this by combining LLM generation with RAG for grounding, iterative feedback for refinement, and robust evaluation metrics to assess quality and originality.

\section{Dataset Collection}
\label{sec:dataset}
We created our dataset by extracting Future Work sentences from 2,354 papers in ACL 2023 and 2,208 papers in ACL 2024, for a total of 4,562 papers. We also collected $1,000$ papers from NeurIPS, along with their open-access peer reviews from OpenReview \footnote{https://openreview.net} from 2021-22. 


 \textbf{1. Author mentioned future work:} We extracted all sections from a paper using the ScienceParse tool \footnote{https://github.com/allenai/science-parse} and extracted the `main review' from OpenReview using Selenium. 
    Then, we constructed the ground truth for future work by extracting content from both the paper and its OpenReview page. Our ground truth includes two components: (1) author-mentioned future work and (2) long-term goals suggested by peer reviewers. The extraction process consisted of two stages—an initial rule-based extraction using a tool, followed by refinement using GPT.
    
    \textbf{A. Tool-Based Extraction}:
    Author-mentioned future work appears in two forms: explicit and implicit.
    For explicit future work, we identified papers with a clearly labeled section such as ``Limitations and Future Work'' and extracted the entire section using the Science Parse tool.
    
    For implicit Future Work, where future directions are discussed within other sections (e.g., Discussion, Conclusion), we searched for the presence of keywords such as “future” or “future work” (case-insensitive). We did not extract from the Abstract, Introduction, Related Work, or Methodology sections, as these typically do not contain future work statements. Once a matching sentence was found, we extracted that sentence and all subsequent ones until the start of the next section. To avoid including unrelated content, parsing was stopped if we encountered any sentence containing keywords like “grant”, “discussion”, or “acknowledgements”. Python regular expressions were used to automate this extraction. We denote the extracted future work by the tool as $F_t$. 
    
    \textbf{B. Re-extracted future work by LLM:} Tool with Regex-based string matching contains various noisy sentences which is not related to future work, showing a high recall, so we further filtered these sentences using LLMs to improve precision. Since most papers do not have a dedicated section, it is often scattered in any other section. Therefore, filtering out noisy sentences is crucial to ensure accurate extraction. Here we employed an \textbf{LLM as an extractor} role to isolate only Future Work sentences while removing irrelevant sentences. We sent the tool-extracted future work $F_t$ to LLM (GPT-4o mini) and prompted it to extract sentences only related to future work, without generating any sentences, and get refined future work $F_g$.  
    
    This produces Future Work paragraphs from $4562$ papers from ACL and $1000$ papers fron NeurIPS, averaging five sentences per paper with an average word length of 63. \\ 

    \textbf{2. OpenReview:} Since no ACL OpenReview were available during our data collection process, we collected OpenReview from the NeurIPS papers only.  
    
    \textbf{A. Tool-Based Extraction:} At first, we used a parsing tool (Selenium) to extract reviews from OpenReview, denoted as $O_{Ft}$. 
    
    \textbf{B. Re-extracted by LLM:} After parsing text from OpenReview, we gathered all peer feedback $O_{Ft}$, and used an LLM to extract potential future‐work suggestions for the authors. But these can include short-term extensions, implementation notes, or minor improvements. To ensure that only meaningful long-term research goals are retained, we applied a second LLM to validate the initially extracted sentences from OpenReview. This step filtered out any content that did not reflect true long-term research directions, resulting in a curated set of goals denoted as $O_{Fg}$. This additional step helps distinguish strategic, forward-looking directions from routine or incremental feedback. 

    \textbf{3. Master Agent:} We employed a master agent LLM to concatenate the author-mentioned future work ($F_g$) with the long-term goals extracted from OpenReview ($O_{Fg}$), without generating any new content. The master agent was designed to identify and merge overlapping or duplicate future work suggestions between $F_g$ and $O_{Fg}$. The entire extraction and merging process was carried out using GPT-4o mini.

\subsection{Evaluation} 
We generated future work with LLM + RAG setting denoted as $F'_G$ where input is Abstract, top 3 sections, and related texts from RAG (Details in Section~\ref{sec:methodology}).
To compare the quality of the tool-extracted future work ($F_t$) and the LLM-extracted future work ($F_g$), we used the LLM-generated future work ($F'_G$) as a reference. Specifically, we conducted an NLP-based evaluation by comparing $F_t$ and $F_g$ against $F'_G$ which is $\text{Eval}_{\text{NLP}}(F_t, F'_G)$ and $\text{Eval}_{\text{NLP}}(F_g, F'_G)$
As shown in Table~\ref{tab:performance-bw-tool-vs-gpt}, the GPT-extracted ground truth ($F_g$) outperforms the tool-extracted ground truth ($F_t$) across nearly all models and metrics, with notable gains in BERTScore, Jaccard Similarity, and Cosine Similarity. The main reason is that $F_t$ often includes noisy or irrelevant sentences, while $F_g$ filters these out and retains only true Future Work content, resulting in stronger alignment with LLM-generated text.



\begin{table}[ht]
  \centering
  \small 
  \begin{tabular}{p{1.80cm} p{0.75cm} p{0.75cm} p{0.75cm} p{0.75cm} p{0.75cm}}
    \toprule
     \textbf{Model} & \textbf{R-L} & \textbf{BScore} & \textbf{JS} & \textbf{CS} & \textbf{Bleu}  \\  
    \midrule 
    \multicolumn{6}{c}{\textbf{Tool extracted Future Work (F\textsubscript{t})}} \\ 
    GPT 3.5 & 16.65   & 86.28  & 13.70  & 44.84  & 3.18  \\    
    GPT4om & 14.22 & 85.91 & 11.55 & 40.53 & 1.29 \\
    GPT4om+RAG &  16.41  & 86.40 & 13.27 & 40.36 & 2.84 \\ 
    \multicolumn{6}{c}{\textbf{GPT Extracted Future Work (F\textsubscript{g})}} \\ 
    GPT 3.5 &  20.64 & 87.88 & 18.49 & 55.36 & 5.61 \\ 
    GPT 4om & 17.69 & 87.44 & 15.87 & 50.52 & 2.72 \\ 
    GPT4om+RAG &  14.19 & 87.67 & 16.95 & 49.32 & 3.42 \\

    \bottomrule 
  \end{tabular} 
  \vspace{3pt}
    \caption{Performance comparison of different models, evaluated against tool- and GPT-extracted future work using LLM-generated future work as the reference, based on ACL papers.
    \footnotesize Note:  GPT 4om, R-L, BScore, JS, CS refers to GPT 4o mini, ROUGE-L, BERTScore, Jaccard Similarity, and Cosine Similarity, respectively.
    }
  \label{tab:performance-bw-tool-vs-gpt}
  \vspace{-18pt}
\end{table}

 
\subsection{Human Evaluation}
\label{sec:hum-eval-ext} 
Additionally, we conducted a human evaluation to assess the effectiveness of LLMs in extracting future work sentences without introducing hallucinated or fabricated content. A user study was carried out on 500 randomly selected samples, involving three annotators who were asked: ``Do you think the LLM extracted future work only, without generating new content or hallucinating?" We provided annotators with both the tool-extracted future work and the LLM-extracted future work for comparison. All annotators agreed that the LLM faithfully extracted existing future work sentences, primarily filtering out noisy content without introducing new information. The annotators were graduate students with a background in machine learning and had no affiliation with this study.

\section{Methodology} 
\label{sec:methodology} 
We propose a multi-stage framework for generating, evaluating, and analyzing LLM-based future work suggestions. The workflow includes: (1) constructing a high-quality dataset, (2) generating future work using LLMs with and without retrieval augmentation (RAG), (3) evaluating the generated outputs against ground truth using both NLP-based and LLM-based metrics, (4) assessing hallucination, novelty, and feasibility, and (5) validating results through human evaluation.

Our work focuses on an LLM-based RAG pipeline that generates future-work sections directly from ACL and NeurIPS papers. At first, we extracted the author's mentioned future work from the paper, then we re-extracted it using GPT. We also did a similar process for OpenReview and made a ground truth containing the author's mentioned future work and long-term goals from OpenReview (Details in ~\ref{sec:dataset}).
In the ACL dataset, the ground truth consists of author-stated future work statements extracted using an LLM, $F_g$. For the NeurIPS dataset, the ground truth includes both the authors’ future work statements and long-term suggestions from OpenReview peer reviewers, ($F_g$ + $O_{Fg}$). In both cases, we constructed input data by collecting all texts from the full paper after removing  `author-mentioned future work'. The future work generated by the LLM with RAG from the input data is denoted as $F'_G$. 

\begin{figure}[ht]
    \centering
    \includegraphics[width=1\linewidth]{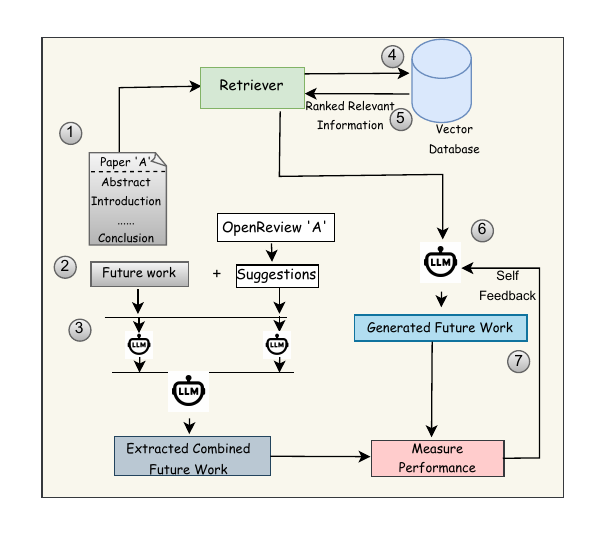}  
    \caption{Overview of our LLM + RAG future-work pipeline: we extract author-written future work from the paper, extract long-term goals from peer reviews via a tool + LLM, generate new future work suggestions with a RAG-augmented LLM from the paper, evaluate their quality, and apply self-feedback for refinement.}
    \label{fig:archi}
    \vspace{-5pt}  
\end{figure}

Figure \ref{fig:archi} depicts our end‐to‐end pipeline: starting from a sample paper, we first extract candidate “future work” sentences $F_t$ via regex with tool and refine (re-extract) them with an LLM and get $F_g$, then pull peer‐review comments from OpenReview and use the same LLM to isolate long-term goals (step 2,3), merging both using master LLM to make a robust ground truth. We remove the author’s mentioned future work, feed the paper to an LLM augmented by a vector-store retriever that supplies related documents, and generate new future work suggestions (step 4,5,6). Each paper is automatically scored on NLP metrics and LLM-based criteria (coherence, relevance, novelty, grammar, overall impression), and any suggestion scoring below a threshold of 3 is critiqued and re-fed into the generator for a polished second iteration (step 7). A more detailed overview of the tasks in this stage is given below.

\textbf{Section Selection:} Processing a full paper has more computational and API costs; to reduce computational and API costs, we experimented with using the abstract and the top three most relevant sections as input to generate future work suggestions. To identify the most relevant sections, we used a cosine similarity-based approach. We calculated the cosine similarity between each section and the paper's  Future Work text (Table \ref{tab:cosine-similarity-performance}) across all papers, and we selected the top three sections with the highest average cosine similarity. These sections formed the basis for generating  Future Work content. For all experiments, we removed the author's mentioned future work from the paper and made the input text.

\textbf{Generating \textit{Future Work} Using LLM and RAG:} Generating future work from a single paper can yield narrow or generic suggestions, often defaulting to boilerplate ideas like ``use more data" or ``apply to new domains.” RAG mitigates this by grounding generation in paper-specific passages and related works, reducing hallucinations and improving relevance. In our experiments, we generated future work using either the top three sections or all sections (excluding the ground-truth future work), supported by an RAG system built from 100 random research papers (Details ~\ref{sec:generation}). The Retriever processes the input query, which includes the prompt and the content of the input paper. It then retrieves additional relevant information from the vector database using a hybrid approach, FAISS + BM25, and applies a ranking mechanism to prioritize the extracted text (Figure~\ref{fig:archi}). Here, BM25 is a traditional term-based method that ranks documents based on keyword overlap and term frequency, and FAISS is a vector-based similarity search library that retrieves texts based on dense embeddings. The final augmented input is fed into the LLM Generator, producing  Future Work content based on the provided context (Figure~\ref{fig:archi}, step 6).

\textbf{LLM as a Judge:} 
Relying solely on NLP metrics (e.g., n-gram overlap, similarity) often miss coherence, accuracy, and fluency. Combining them with LLM-based metrics provides a more holistic evaluation, capturing semantics, logical flow, and human-like judgment while reducing bias toward surface similarity. To assess the quality of generated future work, we used LLM as evaluators. Specifically, we prompted LLMs to rate each suggestion based on six criteria: coherence, relevance, readability, grammar, overall impression, and novelty. Each evaluation was rated on a discrete scale of 1 (worst) to 5 (best), with justifications (Figure \ref{fig:judge}). Our primary evaluator was GPT-4o-mini. To reduce generator–evaluator bias, we also evaluated using LLaMA 3 70B.

\begin{figure}[ht]
    \centering
    \begin{tikzpicture}
        \node[draw, rectangle, rounded corners, fill=yellow!20, text width=0.5\textwidth, inner sep=5pt, scale=0.9] (box) {
            \begin{minipage}{\textwidth}
                \small
                 Instructions: You are provided with two texts for each pair: one is generated by a machine (Machine-Generated Text), and the other is the original or ground truth text (Ground Truth). After reviewing each text, assign a score from 1 to 5 based on the criteria outlined below... 
                Scoring Criteria:
                Coherence and Logic, Relevance and Accuracy. Readability and Style. Grammatical Correctness. Overall Impression. (5: The text is exceptionally coherent; the ideas flow logically and are well connected. 3: The text is coherent but may have occasional lapses in logic or flow. 1: The text is disjointed or frequently illogical.) Task:  
                For each text pair: Rate the Machine-Generated Text on each criterion and provide a final overall score out of 5. Provide a brief justification for your scores, highlighting strengths and weaknesses observed in the machine-generated text relative to the ground truth.
            \end{minipage}
        };
    \end{tikzpicture}
    \caption{Prompt for LLM as a Judge}
    \label{fig:judge}
    \vspace{-5pt}
\end{figure}

\textbf{Iterative Refinement:} 
For each evaluation criterion (coherence, relevance, readability, grammar, overall impression, and novelty), we set a \textit{threshold} of \textbf{3} as an acceptable midpoint (7 for novelty). If an LLM-generated Future Work received a score less than or equal to the midpoint in any metric, the justification was incorporated into the prompt, and the  Future Work was regenerated accordingly (Figure \ref{fig:iterative}). This iterative refinement process was repeated up to two times to assess whether performance improved (Figure \ref{fig:archi}, step 9, 10, and 11). 

\begin{figure}[ht]
    \centering
    \begin{tikzpicture}
        \node[draw, rectangle, rounded corners, fill=yellow!20, text width=0.5\textwidth, inner sep=5pt, scale=0.9] (box) {
            \begin{minipage}{\textwidth}
                \small
                 Your task is to generate a refined "future work" section for a scientific article.
              Input: 
              [Input Paper] \\ 
              Here I am providing the texts and have found these problems. At first, read the feedback and try to improve it when you generate future work. [LLM Feedback] \\
              Based on these details, please generate comprehensive and plausible future work suggestions that could extend the research findings,
              address limitations, and propose new avenues for exploration. Future work should be within 100 words.  
            \end{minipage}
        };
    \end{tikzpicture}
    \caption{Prompt for Iterative Refinement}
    \label{fig:iterative}
    \vspace{-10pt}
\end{figure}

\textbf{Incorporating OpenReview:} 
Authors are often ambiguous or hesitant to disclose ambitious or speculative research plans in their own papers \cite{b11}. Relying solely on author-mentioned Future Work is therefore not ideal, as authors may be reluctant to share their true research directions or may intentionally limit the scope of their suggestions. To address this limitation, we incorporate long-term future work suggestions from OpenReview reviewers and combine them with author-stated future work to construct a more comprehensive ground truth (see Section~\ref{sec:dataset}). These reviewer insights can uncover overlooked or forward-looking directions that are not explicitly stated by the authors but are critical for advancing the field.
We evaluated our approach using three types of ground-truth annotations on NeurIPS data: Author-Mentioned Future Work (FW), OpenReview feedback (OR), and their combination (FW + OR). These labels served as supervision signals to assess the quality of LLM-generated suggestions. We applied this evaluation framework across two model settings: GPT-4o-mini and GPT-4o-mini with RAG. 

\textbf{Measuring Novelty:} 
Measuring novelty is important to assess whether LLM-generated Future Work introduces ideas beyond the ground truth. While alignment ensures relevance, novelty reflects the model’s ability to propose fresh, forward-looking directions. We define novelty as the extent to which the generated text contains ideas absent from the ground truth. Following prior work on LLM-based evaluation, we operationalize novelty as a 0–10 scale judged by an evaluator LLM (GPT-4o mini), where 0 indicates complete overlap (no new ideas) and 10 indicates entirely new research directions. In this scale, a score of around 7 suggests that the output introduces some original ideas while still overlapping with the ground truth, whereas a score of 8 or higher reflects stronger divergence and originality. To evaluate this, we prompted the evaluator LLM to assign a score with justification for each generated output (Figure~\ref{fig:novelty}). Furthermore, if a paper’s novelty score was less than or equal to 7, the justification and input paper were sent back to the LLM to regenerate the Future Work, thereby encouraging more diverse and underexplored directions.


\begin{figure}[ht]
    \centering
    \begin{tikzpicture}
        \node[draw, rectangle, rounded corners, fill=yellow!20, text width=0.5\textwidth, inner sep=5pt, scale=0.9] (box) {
            \begin{minipage}{\textwidth}
                \small
                 You are an expert in evaluating research content for novelty and innovation. I have two sets of text  provided below:
                
                Ground Truth Future Work: 
                
                LLM-Generated Future Work: 
                
                Your task is to compare the LLM-generated future work to the Ground Truth future work and assess its novelty relative to the 
                Ground Truth future work. Follow these steps:
                
                Evaluate Novelty: Identify unique ideas, approaches, or directions in the LLM-generated future work that are not present in 
                the Ground Truth future work. Analyze how innovative or distinct these additions are in the context of the research field.
                
                Quantify Novelty: Provide a novelty score (0-10) for the LLM-generated future work, where 0 indicates complete overlap with the 
                ground truth future work (no new ideas) and 10 indicates entirely new and distinct ideas. Justify the score with a clear reason explaining 
                the extent of novel contributions or lack thereof.
                Present your response in a JSON object with the keys score (integer from 0-10) and reason (a concise explanation of the score). 
            \end{minipage}
        };
    \end{tikzpicture}
    \caption{Prompt for Measuring Novelty}
    \label{fig:novelty}
    \vspace{-15pt}
\end{figure}

\textbf{Measuring Hallucination:} 
Measuring hallucination is crucial in future work generation to ensure that the suggestions are grounded in the original paper’s content and not fabricated or misleading. Hallucinated outputs can misrepresent the study’s scope, overstate limitations, or propose directions that are inconsistent with the paper’s findings. By evaluating hallucination, we can assess the factual consistency and trustworthiness of the generated future work. So we measured hallucination by reframing it as a natural language inference (NLI) task. For each generation, we treated the original paper text concatenated with the ground-truth future work as the premise and the LLM-generated future work as the hypothesis (Figure \ref{fig:hallu}). A judge LLM was employed to ask to classify whether each pair as entailment, neutral, or contradiction, and if the hypothesis is neutral or contradicts the premise, it was marked as hallucination. 

\begin{figure}[ht]
    \centering
    \begin{tikzpicture}
        \node[draw, rectangle, rounded corners, fill=yellow!20, text width=0.5\textwidth, inner sep=5pt, scale=0.9] (box) {
            \begin{minipage}{\textwidth}
                \small
                
                You are a natural language inference (NLI) classifier. 
                Given a premise and hypothesis, respond with exactly one word: 
                ``entailment", ``neutral", or ``contradiction". \\
                Premise: Input paper \\ 
                Hypothesis: LLM generated future work 
            \end{minipage}
        };
    \end{tikzpicture}
    \caption{Prompt for Measuring Hallucination}
    \label{fig:hallu}
    \vspace{-8pt}
\end{figure}

\textbf{Feasibility Check:} 
LLM-generated future work may appear insightful, but it may often lack the grounding needed for actual implementation. If such suggestions are not feasible—given the paper’s methods, data, or research scope—they risk promoting impractical or unrealistic research directions. To address this concern, we conducted feasibility checks on each future work suggestion generated by both the LLM and LLM+RAG settings. Using a separate LLM (GPT-4o-mini) as a judge, we conducted a feasibility assessment by prompting whether each LLM and RAG-generated future work suggestion was executable given the paper's methods, data, or research context (Figure \ref{fig:feas}). We provided an input paper, LLM generated future work to judge LLM, and the judge LLM was instructed to return a binary judgment: feasible or not feasible. This assessment is crucial to ensure that LLM and RAG generated future work is not only meaningful but also applicable and actionable, supporting more grounded and effective research planning. 

\begin{figure}[ht]
    \centering
    \begin{tikzpicture}
        \node[draw, rectangle, rounded corners, fill=yellow!20, text width=0.5\textwidth, inner sep=5pt, scale=0.9] (box) {
            \begin{minipage}{\textwidth}
                \small
                ``You are an expert reviewer. Below is the content of a research paper followed by a suggestion for future work. 
                Evaluate whether the future work is executable in the context of the paper's methodology, dataset, or other components. 
                Respond with exactly one word: 'feasible' or 'not feasible'." \\ 
                Paper Content: Input paper \\ 
                Future Work Suggestion: LLM generated future work 
            \end{minipage}
        };
    \end{tikzpicture}
    \caption{Prompt for Measuring Feasability}
    \label{fig:feas}
    \vspace{-10pt}
\end{figure}


\begin{table}[htbp] 
 \small 
  \centering
  \begin{tabular}{p{0.5cm} p{0.5cm} p{0.5cm} p{0.5cm} p{0.5cm} p{0.5cm} p{0.5cm} p{0.5cm}}
    \toprule
   \textbf{Abs.} & \textbf{Intro.} & \textbf{RW} & \textbf{Data} & \textbf{Meth.} & \textbf{Exp.} & \textbf{Con.} & \textbf{Lim.} \\ 
    \midrule
    \textbf{0.25} & \textbf{0.25} & 0.21 & 0.08 & 0.15 & 0.22 & \textbf{0.22} & 0.21\\
    \bottomrule
  \end{tabular} 
  \vspace{3pt}
  \caption{Average Cosine Similarity of each section with  Future Work. N.B: CS, Abs, Intro, RW, Data, Meth, Exp, Con, Lim means Cosine Similarity, Abstract, Introduction, Releated Work, Data, Methodology, Experiment, Conclusion, and Limitations, respectively.}
  \label{tab:cosine-similarity-performance} 
  \vspace{-15pt}
\end{table}

\begin{table}[htbp]
  \centering
  \small
    \begin{tabular}{c c c}
    \toprule
     \textbf{Metrics} & \textbf{w/o LLM's feed} & \textbf{w LLM's feed}  \\ 
    \midrule
    ROUGE-1 & 24.33 & \textbf{25.74(\scalebox{0.8}{+1.41})} \\  
    ROUGE-2 & 5.27 & \textbf{7.85(\scalebox{0.8}{+1.05})} \\ 
    ROUGE-L & 17.24 & \textbf{20.25(\scalebox{0.8}{+2.58})} \\
    BScore(f1) & 87.23 & \textbf{87.50(\scalebox{0.8}{+0.27})} \\
    Jaccard Sim. & 15.40 & \textbf{18.13(\scalebox{0.8}{+2.73})} \\ 
    Cosine Sim. & 48.07 & \textbf{57.03(\scalebox{0.8}{+8.96})} \\
    BLEU & 2.38 & \textbf{5.74(\scalebox{0.8}{+3.36})} \\
    Coherence & 3.94 & \textbf{3.97(\scalebox{0.8}{+0.03})} \\
    Relevance & 4.07 & \textbf{4.62 (\scalebox{0.8}{+0.55})}\\
    Readability & 3.19 & \textbf{3.49 (\scalebox{0.8}{+0.30})}\\
    Grammar & 4.02 & \textbf{4.01 (\scalebox{0.8}{-0.01})}\\
    Overall & 3.85 & \textbf{3.95 (\scalebox{0.8}{+0.10})} \\ 
 
    \bottomrule
  \end{tabular} 
  \vspace{3pt}
    \caption{Comparison of performance without and with LLM feedback using GPT-4o mini in ACL data.}

  \label{tab:with-without-feedback}
  \vspace{-16pt}
\end{table}

\begin{table}[ht]
\centering
\caption{Future‐Work Suggestions from Different Sources}
\label{tab:future-work-sources}
\begin{tabular}{@{}p{1cm}p{7.5cm}@{}}
\toprule
\textbf{Source} & \textbf{Future‐Work} \\
\midrule
Ground truth & We will explore the application of SWAD (Stochastic Weight
Averaging Densely) to other robustness problems, including ImageNet classification and its robustness benchmarks. \\[0.5em]
LLM          & Future research could explore the applicability of flatness-aware methods in other machine learning contexts, such as reinforcement learning or unsupervised learning. \\[0.5em]
RAG          & Investigate the effectiveness of the SWAD method across a wider variety of datasets and domains, particularly those with significant distribution shifts. This would validate the generalization capabilities of SWAD beyond the current benchmarks. \\
\bottomrule
\end{tabular} 
 \label{tab:gt-llm-rag}
\vspace{-10pt}
\end{table}

\section{Experimental Setup}

\label{sec:generation} 
Our work is future work generation, where we benchmarked a diverse mix of generative and retrieval-augmented methods to understand their relative strengths under realistic constraints. BART and T5 serve as strong, well-studied seq2seq baselines with fixed token-limit trade-offs, while GPT-3.5 and GPT-4o illustrate how off-the-shelf LLMs perform zero-shot under a controlled prompt budget. Fine-tuning LLaMA-3.1 with LoRA/QLoRA and FlashAttention demonstrates that even large open-source models can be adapted efficiently on modest hardware. Finally, integrating RAG grounds generation in concrete evidence, and comparing one-shot versus zero-shot LLM evaluators (GPT vs. Llama) lets us quantify both generation quality and evaluator bias. This multi-axis evaluation ensures our conclusions generalize beyond any single model or configuration. We conducted experiments with LLaMA and its fine-tuning on an A100 GPU with 40 GB VRAM. For BART and T5 models, we utilized Google Colab with a T4 GPU (15 GB VRAM). GPT-4o mini was accessed via the OpenAI API.


\textbf{Generating Similarity between Other Sections and Future Works: } We used Sentence Transformers ('all-MiniLM-L6-v2') for embedding generation and scikit-learn's cosine similarity function for similarity computation.

\textbf{Fine-Tuning Models:} 
First, we fine-tuned BART (1,024-token limit) and T5 (512-token limit) on a 70/30 train/test split, discarding any over-length inputs. 
For LLaMA 3.1 7B fine-tuning, it was trained using `Abstract,' `Introduction,' and `Conclusion' as input, and the extracted ` Future Work' was used as output. Leveraging LLaMA’s extensive pretraining, we fine-tuned it with QLoRA (4-bit, alpha=16) and FlashAttention on a 30/70 train/test split for 60 steps (learning rate 2e-4, 2048-token context). During testing, we provided a prompt with detailed instructions for generating  Future Work, with a maximum output length of 128 tokens and a temperature setting of 1. 
\textbf{Zero/Few Shot(s) LLM:} We ran GPT-3.5 and GPT-4o in zero-shot mode, leveraging their 16 K and 128 K context windows, respectively. For LLM as a judge and LLM as an extractor, we used a zero-shot approach with GPT-4o mini.
\textbf{LLM with RAG Integration:}  We integrated RAG with GPT-4o mini for Future work generation, leveraging OpenAI's `text-embedding-3-small' model for relevant document retrieval. Our vector database comprises 100 randomly selected papers from the dataset, and these papers were removed from the dataset. For semantic search using vector embeddings, we employed LlamaIndex \footnote{https://www.llamaindex.ai/}, utilizing its in-memory vector store rather than an external database. We used a hybrid retriever system to fetch the data from the vector database consisting of BM25 and FAISS with an equal weight of 50\%. 
We segmented the data into chunks of up to 512 tokens to accommodate smaller context windows. The overall context window was set to 3,900 tokens—the maximum number of tokens the LLM can process at a time. Additionally, we set $K=3$, meaning that the top three most relevant chunks are retrieved from the vector store to provide contextual support during the generation process.

\section {Experiments and Results}  
\label{sec:exp-and-results}

\begin{table*}[ht]
  \centering
  \small 
    \begin{tabular}{c c c c c c c c c c c}
    \toprule
     \textbf{Model}  & \textbf{Iteration} & \textbf{R-L} & \textbf{BS} & \textbf{JS} & \textbf{CS} & \textbf{Bleu} & \textbf{Coh} & \textbf{Rel} & \textbf{Read}& \textbf{Gr} \\  
    \midrule 
    GPT 3.5 & 1 &  20.64 & 87.88 & 18.49 & 55.36 & 5.61 & 3.89 & 4.47 & 3.38 & 4.04 \\ 
    Llama 3 Zero Shot & 1 & 10.38  & 85.61 & 14.52 & 52.27 & 2.27 & 3.83 & 4.47 & \textbf{4.04} & 3.44   \\ 
    Llama 3 Fine Tuning & 1 &  - & 85.54 & 12.28 & 45.37 & 1.14 & 3.37 & 3.69 & 2.90 & 3.65 \\ 
    GPT 4o mini & 1 & 17.69 & 87.44 & 15.87 & 50.52 & 2.72 & 3.94 & 4.07 & 3.19 & 4.02 \\ 
    GPT 4o mini + RAG & 1 &  14.19 & 87.67 & 16.95 & 49.32 & 3.42 & 3.93 & 3.96 & 3.18 & 4.01  \\
    GPT 4o mini + RAG (Ours) & 2  &  \textbf{20.87} & \textbf{88.15} & \textbf{18.67} & \textbf{58.33} & \textbf{5.67} &  \textbf{3.97} & \textbf{4.50} & 3.50 & \textbf{4.06}  \\ 
    GPT 4o mini + RAG & 3  & 20.18 & 87.94 & 17.92 & 56.34 & 4.68 & 3.96 & 4.34 & 3.36 & 4.05  \\ 
    \bottomrule 
  \end{tabular}
   \vspace{4pt}
    \caption{Performance comparison of various models in generating  Future Work in ACL data. Input is the top 3 sections with the abstract.
    \footnotesize Note: Metrics include ROUGE (R-L), BERTScore (BS), Cosine Similarity (CS), Jaccard Similarity (JS), BLEU (Bl), Coherence (Coh), Relevance (Rel), Readability (Read), Grammar (Gram). 
    } 
   
  \label{tab:various-models-performance}
  \vspace{-10pt}
\end{table*}

\begin{table*}[htbp]
  \centering 
   \small 
  \begin{tabular}{c c c c c c c c c c c c c}
    \toprule
    \textbf{GT} & \textbf{Iter.} & \textbf{R-L} & \textbf{BS} & \textbf{CS} & \textbf{JS} & \textbf{Bleu} & \textbf{Coh} & \textbf{Rel} & \textbf{Read} & \textbf{Gram} & \textbf{Novelty} & \textbf{Overall} \\ 
    \midrule 
    \multicolumn{13}{c}{\textbf{GPT 4o mini}} \\ 
    FW  & 1 & 15.54 & 85.02 & 68.23 & 11.21 & 2.29 & 4.31 & 4.63 & 3.82 & 4.81 & 7.28 & 4.29  \\ 
    OR  & 1 & 12.34 & 82.94 & 62.56 & 10.43 & 1.26 & 3.67 & 4.37 & 3.19 & 4.18 & 7.84 & 3.65 \\ 
    FW+OR  & 1 & \textbf{18.32} & \textbf{85.20} & \textbf{74.33} & \textbf{14.46} & \textbf{3.55} & 4.41 & \textbf{4.87} & 3.85 & 4.67 & 7.28 & 4.40 \\
    FW+OR  & 2 & 17.43 & 84.63 & 73.50 & 14.29 & 3.42 & 4.48 & 4.79 & 4.15 & 4.89 & \textbf{8.00} & 4.44  \\ \addlinespace
    \multicolumn{13}{c}{\textbf{GPT 4o mini + RAG}} \\ 
    FW  & 1 & 11.89 & 83.00 & 63.27 & 8.72 & 1.08 & 4.26 & 4.04 & 4.87 & \textbf{5.00} & 7.70 & 4.15  \\ 
    OR  & 1 & 12.43 & 82.81 & 62.91 & 10.15 & 1.01 & \textbf{4.56} & 4.56 & \textbf{4.89} & \textbf{5.00} & 7.80 & \textbf{4.52}  \\ 
    FW+OR & 1 & 15.90 & 83.47 & 72.57 & 12.29 & 2.02 & 4.17 & 4.36 & \textbf{4.89} & \textbf{5.00} & 7.11 & 4.38  \\
    \bottomrule
  \end{tabular}
  \vspace{4pt}
  \caption{Performance comparison between different methods in NeurIPS data. Here, GT, FW, OR, Iter means Ground Truth, Author Mentioned Future Work, OpenReview suggestion, and Iterations, respectively.}
  \label{tab:results-neurips} 
  \vspace{-9pt}
\end{table*}


\subsection{Evaluation of Future Work Generation}
We experimented with various LLMs (Table~\ref{tab:various-models-performance}), incorporating RAG to generate Future Work sections. Table \ref{tab:gt-llm-rag} presents qualitative examples of Future Work statements from the original paper (ground truth), LLM-generated output, and RAG-generated output. The RAG-generated examples tend to be longer and contain a greater number of ideas compared to both the ground truth and the LLM-only output 

Traditional NLP-based metrics primarily focused on lexical overlap and semantic quality, and LLM-based evaluation provides more contextual assessment and novelty. NLP-based metrics couldn't measure novelty and relied more on ground truth, missing the depth evaluation. Solely relying on an LLM-based evaluation system raises potential bias issues. To alleviate these problems, we incorporated both LLM-based and NLP-based evaluation systems to make a robust evaluation system for evaluation by focusing on coherence and logic, relevance and accuracy, readability and style, grammatical correctness, overall impression, and novelty. 




\begin{table}[ht]
  \centering
  \small 
  \begin{tabular}{p{2.25cm} p{2.15cm} p{2.05cm}}
    \toprule
     \textbf{Metrics} & \textbf{All sections} & \textbf{3 sections}  \\ 
    \midrule
    ROUGE-1 & \textbf{21.08} & 20.3(\scalebox{0.8}{-0.78})  \\  
    ROUGE-2 & \textbf{6.41} & 5.36(\scalebox{0.8}{-1.05}) \\ 
    ROUGE-L & \textbf{17.03} & 16.65(\scalebox{0.8}{-0.38})  \\
    BScore(f1) & \textbf{86.40} & 86.28(\scalebox{0.8}{-0.12})  \\
    Jaccard S & \textbf{14.54} & 13.70(\scalebox{0.8}{-0.84}) \\ 
    Cosine S & \textbf{45.69} & 44.84(\scalebox{0.8}{-0.85}) \\
    BLEU & \textbf{4.13} & 3.18(\scalebox{0.8}{-0.95})   \\
    \bottomrule
  \end{tabular} 
  \vspace{4pt}
  \caption{Performance comparison of GPT-3.5 in generating  Future Work using three selected sections versus full-text input, evaluated against both Tool extracted Ground Truth (GT) and GPT-extracted ground truth on ACL data.}
  \label{tab:3sec-vs-allsec} 
  
  \vspace{-19pt}
\end{table}

\textbf{Evaluation-among all models:} We evaluated multiple models and found that GPT-4o mini with RAG performed best overall when LLM feedback was incorporated once (iteration 2), as shown in Table~\ref{tab:various-models-performance}. This configuration achieved the highest scores across both traditional NLP metrics and LLM-based evaluations.
Interestingly, applying RAG with GPT-4o mini in the first iteration slightly reduced performance compared to using GPT-4o mini alone. For example, ROUGE-L dropped by 3.5 points and cosine similarity by 1.2 points, while coherence, relevance, readability, and grammar also saw small decreases. This suggests that introducing external knowledge too early may sometimes add noise or distract the model from the original paper content.
For the NeurIPS dataset (Table~\ref{tab:results-neurips}), RAG did not improve NLP-based metrics when evaluated against author-mentioned future work (FW), OpenReview reviews (OR), or their combination (FW+OR). However, it did yield modest gains on LLM-based evaluations. For example, when FW was used as ground truth, readability improved by +1.05 and grammar by +0.19; and when FW+OR was used, readability improved by +1.04 and grammar by +0.33. These results highlight that while RAG may not always boost traditional overlap metrics, it can enhance the fluency and readability of generated text.

\textbf{Evaluation-incorporating LLM feedback:} After generating the Future Work text, we used a separate LLM as a judge, providing scores and short justifications. This feedback was then fed back into the generation process to refine the output. Our experiments show that adding feedback once significantly improved performance across multiple evaluation metrics (Table~\ref{tab:with-without-feedback}). For instance, a single feedback round helped produce text that was more accurate, coherent, and aligned with the source paper.
Interestingly, we observed that one round of feedback was optimal. While the first feedback loop (iteration 2) boosted performance for GPT-4o mini with RAG (Table~\ref{tab:various-models-performance}), a second round (iteration 3) actually reduced performance, likely because repeated feedback introduced bias.
On the NeurIPS dataset (Table~\ref{tab:results-neurips}), the same pattern emerged: a second feedback round provided small improvements in readability and grammar but slightly lowered overlap-based metrics. Since GPT-4o mini + RAG already achieved strong scores after one iteration, additional feedback was unnecessary. Overall, these findings highlight that a single round of feedback integration balances improvement with stability, making it the most effective strategy. 

\textbf{Novelty Evaluation:} We evaluated the novelty of LLM- and RAG-generated Future Work using three ground-truth settings: author-mentioned future work (FW), OpenReview reviews (OR), and their combination (FW + OR) (Table~\ref{tab:results-neurips}). Unlike established metrics such as BLEU or precision, novelty scoring is inherently subjective, as it depends on the evaluator model and the chosen ground truth. We therefore interpret novelty comparatively across settings rather than as an absolute value. Across all ground truths, the relative trends were consistent: GPT-4o mini produced reasonably novel suggestions, and adding RAG slightly increased novelty when FW was used as the reference (+0.42), though it led to a small decrease when FW + OR was used (-0.17). The most notable improvement came from introducing self-feedback, which raised the novelty score from 7.28 to 8.00, the highest across all settings. This indicates that a single round of feedback helps the model generate more diverse and underexplored research directions, while the comparative analysis across FW, OR, and FW+OR provides a more robust view of originality.

\textbf{LLM as a Judge:} All of our evaluation experiments—including future work extraction and the assessment of coherence, relevance, readability, grammar, novelty, hallucination, and feasibility—were conducted using GPT-4o mini. While LLM-based evaluation is inherently more subjective than standard metrics such as BLEU or ROUGE, prior work has shown that LLMs can approximate human judgments with reasonable reliability. In our case, GPT-4o mini produced evaluations that aligned well with human annotations (see Section~\ref{sec:exp-and-results}, B). To check robustness, we also experimented with LLaMA 3 70B, which produced less reliable justifications and tended to inflate scores even for low-quality outputs. These differences highlight that absolute values should be interpreted cautiously, and our results are best understood in comparative terms across settings (e.g., LLM-only vs. RAG, with vs. without feedback) rather than as fixed absolute scores.



\textbf{Hallucination Rate:} To assess whether LLM- and RAG-generated Future Work contained hallucinations, we employed a natural language inference (NLI)-based approach that treated the input paper as the reference. Hallucination rates were then evaluated using two judge models: GPT-4o mini and LLaMA 3 8B (Table~\ref{tab:hallucination-rate}). When GPT-4o mini served as the judge, about one in four generated Future Work statements (26.26\%) contained information not supported by the source paper or its ground-truth future work, but this rate dropped to 19.52\% when RAG was incorporated, indicating that retrieval helps ground the generated text more closely in the source content. In contrast, LLaMA 3 8B judged the same outputs much more conservatively, reporting only 3.60\% hallucinations. However, this very low rate appears to reflect the limitations of LLaMA 3 8B as a judge rather than stronger grounding: the model has a shorter context window (8K) and fewer parameters (8B), and in practice, it almost always classified the LLM-generated text as entailment. For this reason, we do not rely on LLaMA 3 8B as a reliable judge. Overall, hallucination rates should be interpreted comparatively rather than absolutely: the key finding is that RAG consistently reduces hallucination relative to LLM-only generation.

\textbf{Feasibility Check:} We measure how feasible LLM + RAG-generated future work is in terms of data and methodology. We employed a new GPT-4o mini as evaluator, and the evaluation revealed that 98.92\% of the LLM + RAG-generated future work was classified as feasible, indicating strong contextual alignment between the proposed future work and the methodology and data described in the source papers.




\begin{table}[htbp] 
  \small 
  \centering
  \begin{tabular}{p{2.52cm} p{2.52cm} p{2.50cm}}
    \toprule
     \textbf{Text Generation} & \textbf{Evaluator} & \textbf{Hallucination Rate} \\ 
    \midrule
    GPT 4o mini & GPT 4o mini & 26.26 \%\\
    GPT 4o mini+RAG & GPT 4o mini & 6.74 \%\\ 
    GPT 4o mini & Llama 3 8B & 3.60 \%\\
    \bottomrule
  \end{tabular} 
   \vspace{3pt}
  \caption{Hallucination rate of each model in NeurIPS data (lower is better)}
  \label{tab:hallucination-rate} 
  \vspace{-13pt}
\end{table}


\begin{table}[htbp]
  \small
  \centering
  \begin{tabular}{p{1.2cm}p{1.3cm}p{1.3cm}p{1.3cm}}
    \hline
   \textbf{Ques.} & \textbf{Rating (Avg.)} & \textbf{W. Kappa} & \textbf{Kend' Tau} \\ \hline
    Q1 & 2.12 &  \textbf{0.30} & \textbf{0.36} \\ 
    Q2 & \textbf{2.75} &  0.28 & 0.29 \\  \hline
  \end{tabular} 
   \vspace{3pt}
  \caption{Average rating and annotators user agreement on user study.}
  \label{tab:user-agreement}
   \vspace{-18pt} 
\end{table}

\subsection{Human Evaluation}
\label{sec:human-eval} 

Using the same LLM (GPT-4o mini) as the extractor, generator, and evaluator may introduce potential biases, such as self-validation and confirmation bias. To mitigate this concern, we conducted a threefold human evaluation who are PhD-level NLP researchers (not affiliated with this study). \textbf{1. Evaluating the Extractor (LLM as Extractor):} We asked three independent annotators to assess whether GPT-4o mini accurately extracted future work statements without generating new content or hallucinating. All annotators confirmed that the LLM faithfully extracted relevant content without introducing noise or fabricated text (see Section~\ref{sec:dataset}, B). \textbf{2. Evaluating the Generator (LLM as Generator):} To assess the quality of the LLM-generated future work, we randomly sampled 100 examples and asked three annotators to rate the generated outputs. The evaluation focused on the question:
Q1: How well does the LLM-generated future work align with the ground truth? We provided annotators with the LLM-extracted ground truth alongside the LLM-generated future work for evaluation. As shown in Table~\ref{tab:user-agreement}, the responses yielded a moderate average score of 2.12 out of 3, with a Cohen's Kappa of 0.30 and Kendall's Tau of 0.36, indicating consistent agreement among annotators. \textbf{3. Evaluating Feedback (LLM as Feedback Provider):} To evaluate the effect of LLM-based self-feedback, we asked annotators the following:
Q2: Does the LLM feedback iteration improve the originality and quality of the generated future work? 
We provided annotators with the initial LLM-generated future work alongside its revised version after incorporating one round of LLM feedback. The one-round feedback approach achieved a high average score of 2.75 out of 3, suggesting that the feedback loop significantly enhanced the originality and overall usefulness of the generated content.

\subsection{Ablation Study} 
\textbf{Evaluation-considering all sections vs top-3 sections.} 
We used cosine similarity to identify the three most relevant sections in the ACL dataset. Future work was then generated using two input settings: (1) the top three most relevant sections, and (2) all sections excluding the author-mentioned future work. We evaluated performance using NLP-based metrics by comparing the LLM-generated future work against the ground truth (author-mentioned future work combined with OpenReview suggestions). As shown in Table~\ref{tab:3sec-vs-allsec}, using only the top three sections led to a slight performance drop across all metrics, including ROUGE-L (-0.78), BERTScore (-0.12), Jaccard Similarity (-0.84), and Cosine Similarity (-0.85).

\textbf{Evaluation incorporating OpenReview} We conducted an ablation study using three types of ground truth: FW (author-mentioned future work), OR (long-term goals from OpenReview), and FW + OR (a combination of both). As shown in Table~\ref{tab:results-neurips}, the combined FW + OR setting consistently produced the best results. For example, when GPT-4o mini was used as the generator, FW + OR outperformed FW alone with clear gains in overlap and similarity metrics (e.g., ROUGE-L improved by +2.78 and cosine similarity by +6.10). Compared to OR alone, the combined setting performed even better, with ROUGE-L increasing by nearly +6 points and cosine similarity by almost +12 points. These improvements suggest that combining author-written and peer-reviewed goals gives the model a broader and more balanced reference, leading to richer and more diverse future work suggestions.

A similar pattern was observed when GPT-4o mini + RAG was used as the generator. The combined FW + OR ground truth again outperformed FW alone on most metrics, including a +4.01 gain in ROUGE-L and a +9.30 gain in cosine similarity. Compared to OR alone, the combined setting also showed consistent advantages, though some LLM-based evaluation metrics saw minor declines. Overall, the results indicate that merging author and reviewer perspectives creates a stronger ground truth, allowing LLMs to generate future work that better captures the diversity of research directions.

GPT-extracted Ground Truth aligns well with LLM-generated  Future Work text than the tool-extracted ground truth, indicating that the tool-extracted Ground Truth contains more noise and irrelevant sentences.

\section{Discussions} 
Our experiments focused on NLP and ML papers (ACL and NeurIPS) because these venues provide large, open-access datasets with rich peer-review feedback, making them ideal for benchmarking. However, the framework itself is not domain-specific. Extending it to other areas such as biomedicine or social sciences would primarily require (1) constructing domain-specific retrieval corpora, (2) curating ground truths from available peer reviews or domain experts, and (3) adjusting evaluation criteria to reflect community-specific standards. While the core methodology—RAG for grounding, LLM-as-a-judge evaluation, and iterative self-feedback—remains the same, the effort lies mainly in dataset preparation and defining appropriate evaluation references for each field.
\textbf{RQ1: How reliable is an LLM in extracting future work
statements compared to traditional tool-based methods?} Traditional tools often miss Future Work when it is not a distinct section, whereas our LLM-based extractor can identify relevant content in such cases. Using LLM-extracted text as reference improves performance across NLP- and LLM-based metrics compared to tool-based extraction, as it filters out noise and yields cleaner ground truth. Human evaluations further confirm its reliability, showing that the LLM extracts relevant content without hallucination. \textbf{RQ2: How does input context selection (top-3 vs. all sections) impact the quality of LLM-generated future work, and how does RAG influence these results?} Using only the top three most relevant sections, instead of the full paper, resulted in a slight decline across most NLP-based evaluation metrics. However, incorporating RAG with GPT-4o mini improved overall performance, reduced hallucinations, and increased some LLM based metrics, but it decreased the performance in n-gram or semantic similarity metrics as new information comes from the vector database. \textbf{RQ3: How does expanding ground truth with peer-review goals and incorporating LLM feedback impact the performance and quality of LLM-generated future work?} Augmenting the ground truth with long-term goals extracted from OpenReview peer reviews led to consistent improvements across all evaluation metrics. Which depicts LLM-generated future work can be more effectively grounded when both the author-mentioned future work and OpenReview feedback are used together as reference, rather than relying on a single source alone. Furthermore, a single round of LLM self-feedback consistently improved performance across both NLP-based and LLM-based evaluations. However, adding a second round of feedback introduced verbosity and reduced overall performance, suggesting diminishing returns with excessive iteration. 
Throughout our study, GPT-4o mini served as both the extractor and evaluator, demonstrating strong alignment with human judgments.

\section{Conclusions}
The Future Work section is a forward-looking guide, helping the research community explore new directions. We utilized an LLM to extract Future Work, producing a more coherent ground truth that enhances model performance and incorporated a strong ground truth from OpenReview to provide more broader perspective. Additionally, we integrated an external vector database to further improve LLM's performance to generate Future Work from input text. For evaluation, we applied NLP-based metrics alongside an LLM-as-a-Judge approach, using explainable LLM metrics to assess performance, provide feedback, and iteratively refine text generation. One-time feedback improves the performance in all NLP and LLM-based metrics. Moreover, we measure hallucination rate, feasibility, and novelty to ensure that the generated future work is grounded in the paper’s content, realistically achievable given the methods and data, and offers original, valuable directions beyond what is already stated. 





\section*{Limitations and  Future Work} 
Our analysis is limited to ACL papers (2012–2024) and NeurIPS (2021–22), ensuring domain relevance but restricting cross-disciplinary generality. Using a single model for iterative refinement also risks stylistic convergence. For baselines (T5, BART), we capped inputs at 512–1,024 tokens and did not explore advanced feedback strategies (e.g., chain-of-thought, self-consistency). Our random selection of 100 papers for RAG may include works published after the target paper; in future, we will restrict retrieval to prior publications. Novelty was assessed using an LLM judge, but we plan to incorporate human evaluation for greater reliability. Looking ahead, we aim to extend our pipeline to other domains, evaluate open-source LLMs to reduce costs, build domain-specific retrieval corpora, and create a gold-standard dataset with more human annotators. We also plan to integrate cited-by literature, advanced reasoning techniques, and evaluator alignment methods (RLHF/RLAIF), while releasing an open-source tool that generates Future Work suggestions directly from uploaded PDFs.

\section*{Ethics Statement} 
We recognize ethical considerations in extracting and generating Future Work—including intellectual property, authorship, and responsible AI use. Our framework is strictly assistive: it acknowledges sources, avoids claims of ownership, and anchors suggestions in the original paper through RAG, iterative refinement, and human feedback. It does not replace human insight but helps researchers organize and clarify directions. Still, we acknowledge risks: misuse could foster over-reliance on AI and weaken critical reflection. As generative models advance toward broader sections of scientific writing, safeguards, transparency, and human responsibility remain essential. Our aim is to support—not supplant—authorship while preserving integrity and enabling more robust development of future research directions.

\section*{Acknowledgments}
We acknowledge the use of ChatGPT to refine the writing of this paper, specifically to improve readability, clarity, and flow. The research ideas, experimental design, data collection, analyses, and substantive contributions are entirely the work of the authors. ChatGPT was used only as a language assistant to polish text expression, and not for generating content, research ideas, or results.


\vspace{12pt}


\begin{thebibliography}{00} 
 
\bibitem{b1} Nicholas, David, et al. ``Do younger researchers assess trustworthiness differently when deciding what to read and cite and where to publish?." International Journal of Knowledge Content Development \& Technology 5.2 (2015).

\bibitem{b2} Aguinis, Herman, Ravi S. Ramani, and Nawaf Alabduljader. ``What you see is what you get? Enhancing methodological transparency in management research." Academy of Management Annals 12.1 (2018): 83-110. 





\bibitem{b4} Hara, Noriko, et al. ``An emerging view of scientific collaboration: Scientists' perspectives on collaboration and factors that impact collaboration." Journal of the American Society for Information science and Technology 54.10 (2003): 952-965.

\bibitem{b5} Hyder, Adnan A., et al. ``National policy-makers speak out: are researchers giving them what they need?." Health policy and planning 26.1 (2011): 73-82. 

\bibitem{b6} Thelwall, Mike, et al. ``What is research funding, how does it influence research, and how is it recorded? Key dimensions of variation." Scientometrics 128.11 (2023): 6085-6106.


\bibitem{b8} Kelly, Jacalyn, Tara Sadeghieh, and Khosrow Adeli. ``Peer review in scientific publications: benefits, critiques, \& a survival guide." Ejifcc 25.3 (2014): 227. 

\bibitem{b9} Conaway, Carrie, Venessa Keesler, and Nathaniel Schwartz. ``What research do state education agencies really need? The promise and limitations of state longitudinal data systems." Educational Evaluation and Policy Analysis 37.1suppl (2015): 16S-28S.

\bibitem{b10} Ortagus, Justin C., et al. ``Performance-based funding in American higher education: A systematic synthesis of the intended and unintended consequences." Educational Evaluation and Policy Analysis 42.4 (2020): 520-550.

\bibitem{b11} Suray, Jacques, et al. ``How the  Future Works at SOUPS: Analyzing  Future Work Statements and Their Impact on Usable Security and Privacy Research." arXiv preprint arXiv:2405.20785 (2024). 

\bibitem{b12} Wang, Hanchen, et al. ``Scientific discovery in the age of artificial intelligence." Nature 620.7972 (2023): 47-60.

\bibitem{b13} Si, Chenglei, Diyi Yang, and Tatsunori Hashimoto. ``Can llms generate novel research ideas? a large-scale human study with 100+ nlp researchers." arXiv preprint arXiv:2409.04109 (2024).

\bibitem{b14} Ashkinaze, Joshua, et al. ``How AI Ideas Affect the Creativity, Diversity, and Evolution of Human Ideas: Evidence from a Large, Dynamic Experiment." arXiv preprint arXiv:2401.13481, 2024.

\bibitem{b15} Anderson, Barrett R., et al. ``Homogenization Effects of Large Language Models on Human Creative Ideation." Proceedings of the 16th Conference on Creativity \& Cognition, 2024, pp. 413–25.

\bibitem{b16} Lin, Chin-Yew. ``ROUGE: A Package for Automatic Evaluation of Summaries." Text Summarization Branches Out, 2004, pp. 74–81.

\bibitem{b17} Papineni, Kishore, et al. ``BLEU: A Method for Automatic Evaluation of Machine Translation." Proceedings of the 40th Annual Meeting of the Association for Computational Linguistics, 2002, pp. 311–18.

\bibitem{b18} Zhang, Tianyi, et al. ``BERTScore: Evaluating Text Generation with BERT." arXiv preprint arXiv:1904.09675, 2019.

\bibitem{b19} Grootendorst, Maarten. ``BERTopic: Neural Topic Modeling with a Class-Based TF-IDF Procedure." arXiv preprint arXiv:2203.05794, 2022.

\bibitem{b20} Gonçalves, Sérgio, et al. ``A Deep Learning Approach for Sentence Classification of Scientific Abstracts." Artificial Neural Networks and Machine Learning—ICANN 2018, edited by [Editor], Springer, 2018, pp. 479–88.

\bibitem{b21} Houngbo, Hospice, and Robert E. Mercer. ``Method Mention Extraction from Scientific Research Papers." Proceedings of COLING 2012, 2012, pp. 1211–22.

\bibitem{b22} Al Azher, Ibrahim, et al. ``LimTopic: LLM-Based Topic Modeling and Text Summarization for Analyzing Scientific Articles Limitations." 2024 ACM/IEEE Joint Conference on Digital Libraries (JCDL), 2024.

\bibitem{b23} Azher, Ibrahim Al. ``Generating Suggestive Limitations from Research Articles Using LLM and Graph-Based Approach." Proceedings of the 24th ACM/IEEE Joint Conference on Digital Libraries, 2024, pp. 1–3.

\bibitem{b24} Al Azher, Ibrahim, and Hamed Alhoori. ``Mitigating Visual Limitations of Research Papers." 2024 IEEE International Conference on Big Data (BigData), 2024, pp. 8614–16.

\bibitem{b25} Hu, Yue, and Xiaojun Wan. ``Mining and Analyzing the  Future Works in Scientific Articles." arXiv preprint arXiv:1507.02140, 2015.

\bibitem{b26} Zhang, Chengzhi, et al. ``Automatic Recognition and Classification of  Future Work Sentences from Academic Articles in a Specific Domain." Journal of Informetrics, vol. 17, no. 1, 2023, p. 101373.

\bibitem{b27} Song, Ruoxuan, et al. ``Identifying Academic Creative Concept Topics Based on  Future Work of Scientific Papers." Data Analysis and Knowledge Discovery, vol. 5, no. 5, 2021, pp. 10–20.

\bibitem{b28} Hao, Wenke, et al. ``The ACL FWS-RC: A Dataset for Recognition and Classification of Sentences about  Future Works." Proceedings of the ACM/IEEE Joint Conference on Digital Libraries in 2020, 2020, pp. 261–69.

\bibitem{b29} Qian, Yuchen, et al. ``Using  Future Work Sentences to Explore Research Trends of Different Tasks in a Special Domain." Proceedings of the Association for Information Science and Technology, vol. 58, no. 1, 2021, pp. 532–36.

\bibitem{b30} Zhu, Zihe, et al. ``Recognizing Sentences Concerning Future Research from the Full Text of JASIST." Proceedings of the Association for Information Science and Technology, vol. 56, no. 1, 2019, pp. 858–59.

\bibitem{b31} Suray, Jacques, et al. ``How the  Future Works at SOUPS: Analyzing  Future Work Statements and Their Impact on Usable Security and Privacy Research." arXiv preprint arXiv:2405.20785, 2024.

\bibitem{b32} Radensky, Marissa, et al. ``Scideator: Human-LLM Scientific Idea Generation Grounded in Research-Paper Facet Recombination." arXiv preprint arXiv:2409.14634, 2024.

\bibitem{b33} Qi, Biqing, et al. ``Large Language Models Are Zero Shot Hypothesis Proposers." arXiv preprint arXiv:2311.05965, 2023.

\bibitem{b34} Manning, Benjamin S., et al. ``Automated Social Science: Language Models as Scientist and Subjects. National Bureau of Economic Research, 2024.

\bibitem{b35} Jain, Moksh, et al. ``GFlowNets for AI-Driven Scientific Discovery." Digital Discovery, vol. 2, no. 3, 2023, pp. 557–77.

\bibitem{b36} Lu, Chris, et al. ``The AI Scientist: Towards Fully Automated Open-Ended Scientific Discovery." arXiv preprint arXiv:2408.06292, 2024.

\bibitem{b37} Si, Chenglei, et al. ``Can LLMs Generate Novel Research Ideas? A Large-Scale Human Study with 100+ NLP Researchers." arXiv preprint arXiv:2409.04109, 2024.

\bibitem{b38} Ouyang, Long, et al. ``Training Language Models to Follow Instructions with Human Feedback." Advances in Neural Information Processing Systems, vol. 35, 2022, pp. 27730–44.

\bibitem{b39} Chen, Xinyun, et al. ``Teaching Large Language Models to Self-Debug." arXiv preprint arXiv:2304.05128, 2023.

\bibitem{b40} Wang, Xuezhi, et al. ``Self-Consistency Improves Chain of Thought Reasoning in Language Models." arXiv preprint arXiv:2203.11171, 2022.

\bibitem{b41} Fu, Yao, et al. ``Improving Language Model Negotiation with Self-Play and In-Context Learning from AI Feedback." arXiv preprint arXiv:2305.10142, 2023.

\bibitem{b42} Yang, Zonglin, et al. ``Large Language Models for Automated Open-Domain Scientific Hypotheses Discovery." arXiv preprint arXiv:2309.02726, 2023.

\bibitem{b43} Peng, Baolin, et al. ``Check Your Facts and Try Again: Improving Large Language Models with External Knowledge and Automated Feedback." arXiv preprint arXiv:2302.12813, 2023.

\bibitem{b44} Madaan, Aman, et al. ``Self-Refine: Iterative Refinement with Self-Feedback." Advances in Neural Information Processing Systems, vol. 36, 2023, pp. 46534–94.

\bibitem{b45} Chiang, Cheng-Han, and Hung-yi Lee. ``Can Large Language Models Be an Alternative to Human Evaluations?" arXiv preprint arXiv:2305.01937, 2023.

\bibitem{b46} Nguyen, Huyen, et al. ``A Comparative Study of Quality Evaluation Methods for Text Summarization." arXiv preprint arXiv:2407.00747, 2024.

\bibitem{b47} Hu, Edward J., et al. ``LoRA: Low-Rank Adaptation of Large Language Models." ICLR, vol. 1, no. 2, 2022, p. 3.

\bibitem{b48} Dettmers, Tim, et al. ``QLoRA: Efficient Finetuning of Quantized LLMs." Advances in Neural Information Processing Systems, vol. 36, 2023, pp. 10088–115.

\bibitem{b49} Dao, Tri, et al. ``FlashAttention: Fast and Memory-Efficient Exact Attention with IO-Awareness." Advances in Neural Information Processing Systems, vol. 35, 2022, pp. 16344–59.

\bibitem{b50} Teufel, Simone. ``Do"  Future Work" sections have a purpose? Citation links and entailment for global scientometric questions." BIRNDL@ SIGIR (1). 2017.

\bibitem{b51} Qian, Yuchen, et al. ``Using  Future Work sentences to explore research trends of different tasks in a special domain." Proceedings of the Association for Information Science and Technology 58.1 (2021): 532-536.

\bibitem{b52} Lewis, Patrick, et al. ``Retrieval-augmented generation for knowledge-intensive nlp tasks." Advances in neural information processing systems 33 (2020): 9459-9474. 

\bibitem{b53} Azher, Ibrahim Al, et al. ``BAGELS: Benchmarking the Automated Generation and Extraction of Limitations from Scholarly Text." arXiv preprint arXiv:2505.18207 (2025).





































































\end{thebibliography}
\end{document}